\pdfoutput=1

\documentclass[11pt]{article}

\usepackage{acl} 

\usepackage{times}
\usepackage{latexsym}

\usepackage[T1]{fontenc}

\usepackage[utf8]{inputenc}

\usepackage{microtype}

\usepackage{graphicx}
\usepackage{subcaption}

\usepackage{algorithm}  
\usepackage{algorithmicx} 
\usepackage{algpseudocode}

\usepackage{multirow}
\usepackage{booktabs}
\usepackage{colortbl} 
\usepackage{arydshln} 

\usepackage{amsmath}
\usepackage{amssymb}
\usepackage{verbatim}
\usepackage{siunitx}
\usepackage{bm}

\usepackage{appendix}

\title{Hierarchical Inductive Transfer for Continual Dialogue Learning}

\author{Shaoxiong Feng$^{\text{1,2}}$\enskip Xuancheng Ren$^{\text{3}}$\enskip Kan Li$^{\text{1}}$\enskip Xu Sun$^{\text{3,4}}$\\
$^{\text{1}}$Beijing Institute of Technology $^{\text{2}}$University of Technology Sydney\\ 
$^{\text{3}}$MOE Key Laboratory of Computational Linguistics, School of CS, Peking University\\ 
$^{\text{4}}$Beijing Academy of Artificial Intelligence\\ 
{\small \tt \{shaoxiongfeng, likan\}@bit.edu.cn \{renxc, xusun\}@pku.edu.cn}}

\begin{document}
\maketitle

\begin{abstract}
Pre-trained models have achieved excellent performance on the dialogue task. 
However, for the continual increase of online chit-chat scenarios, directly fine-tuning these models for each of the new tasks not only explodes the capacity of the dialogue system on the embedded devices but also causes knowledge forgetting on pre-trained models and knowledge interference among diverse dialogue tasks. 
In this work, we propose a hierarchical inductive transfer framework to learn and deploy the dialogue skills continually and efficiently. 
First, we introduce the adapter module into pre-trained models for learning new dialogue tasks. As the only trainable module, it is beneficial for the dialogue system on the embedded devices to acquire new dialogue skills with negligible additional parameters. 
Then, for alleviating knowledge interference between tasks yet benefiting the regularization between them, we further design hierarchical inductive transfer that enables new tasks to use general knowledge in the base adapter without being misled by diverse knowledge in task-specific adapters. 
Empirical evaluation and analysis indicate that our framework obtains comparable performance under deployment-friendly model capacity.
\end{abstract}

\section{Introduction}

Neural dialogue models \cite{DBLP:conf/acl/ShangLL15,DBLP:conf/aaai/SerbanSBCP16,DBLP:conf/naacl/LiGBGD16} have drawn increasing attention due to their high commercial value. 
Previous work usually makes efforts to improve the diversity and coherence of the responses \cite{DBLP:conf/aaai/SerbanSLCPCB17,DBLP:conf/ijcai/ZhangLGXC18,DBLP:conf/nips/ZhangGGGLBD18,DBLP:conf/emnlp/FengRCSLS20,DBLP:conf/acl/SunFLLL20}. 
However, the application of neural dialogue models also requires advanced conversation skills, and 
recently, a lot of work tries to enable models to express empathy \cite{DBLP:conf/aaai/ZhouHZZL18,DBLP:conf/acl/RashkinSLB19}, be knowledgeable \cite{DBLP:conf/aaai/GhazvininejadBC18,DBLP:conf/iclr/DinanRSFAW19}, and demonstrate consistent personalities \cite{DBLP:conf/ijcai/QianHZXZ18,DBLP:conf/acl/KielaWZDUS18,DBLP:journals/www/ZhangZWZL19}. 
Specifically, the dialogue model is trained on a task-specific dataset to learn the corresponding conversation skill. 
However, with the increasing number of online chit-chat scenarios, the dialogue system is further expected to continually specialize in new tasks without sacrificing the performance on old tasks. 
Meanwhile, the dialogue system must keep its capacity as small as possible for the deployment on the computation resource--limited embedded devices.

Pre-trained models \cite{radford2018improving,DBLP:conf/naacl/DevlinCLT19} have successfully facilitated the learning of the downstream tasks in various fields. 
To address the challenge of continual dialogue learning, directly fine-tuning pre-trained models on each of the new dialogue tasks is a straightforward way to equip the dialogue system with new conversation skills continually. However, it explodes the capacity of the dialogue system because knowledge of new tasks need to be stored in new pre-trained models to avoid erasing knowledge of old tasks. 
A more advanced approach is to multi-task one pre-trained model on all old tasks and then fine-tune it on new tasks, which can alleviate the capacity problem and use general knowledge between old tasks to improve the model performance on new tasks \cite{DBLP:conf/acl/SmithWSWB20}. 
Nonetheless, these advantages come at the cost of performance decline on some old tasks due to knowledge interference between diverse tasks.

To tackle these problems, we propose a hierarchical inductive transfer framework to construct and deploy the dialogue system with fewer computational resources. 
The framework is inspired by the fact that the conversational skills are multi-layered, and while general skills, e.g., uttering fluent sentences, are necessary for all scenarios and the requisite for sophisticated skills, specialized skills, such as negotiating and debating, work for fewer occasions. 
In the hierarchy of conversational skills, the latter skills could be efficiently built upon the former skills if they are well-learned.
However, considering it is difficult to determine the proper order of the skills and the skills needed for a dataset, we take the following practical approach.

We first introduce the adapter module, consisting of a small sub-net, into the pre-trained model. 
Each block of the pre-trained model is assigned two adapters inserted after the self-attention layer and the feed-forward layer. 
During training, adapters, as the only trainable parameters, learn knowledge of dialogue tasks, which avoids knowledge forgetting on pre-trained models and therefore keeps the capacity of the dialogue system almost constant as the number of dialogue tasks increases. 
Then, we separate the adapter into the base adapter and the task-specific adapter to avoid the performance decline of models on old tasks caused by knowledge interference between diverse tasks. 
The former is multi-tasked with old tasks to obtain general knowledge by regularization between diverse tasks, which facilitates the learning of new tasks. The latter is further fine-tuned on any dialogue task to learn the corresponding task-specific knowledge, which maintains the model performance on old tasks. 
Finally, the proposed framework significantly enhances the training efficiency due to the learning of dialogue tasks only being conducted via adapters.
\section{Method}

In this section, we first describe the vanilla adapter and how to apply it to the dialogue tasks and then present the hierarchical inductive transfer to learn general knowledge and task-specific knowledge.

\subsection{Adapter for Continual Dialogue Learning}
\begin{figure}[t]
    \centering
    \includegraphics[width=1.0\linewidth]{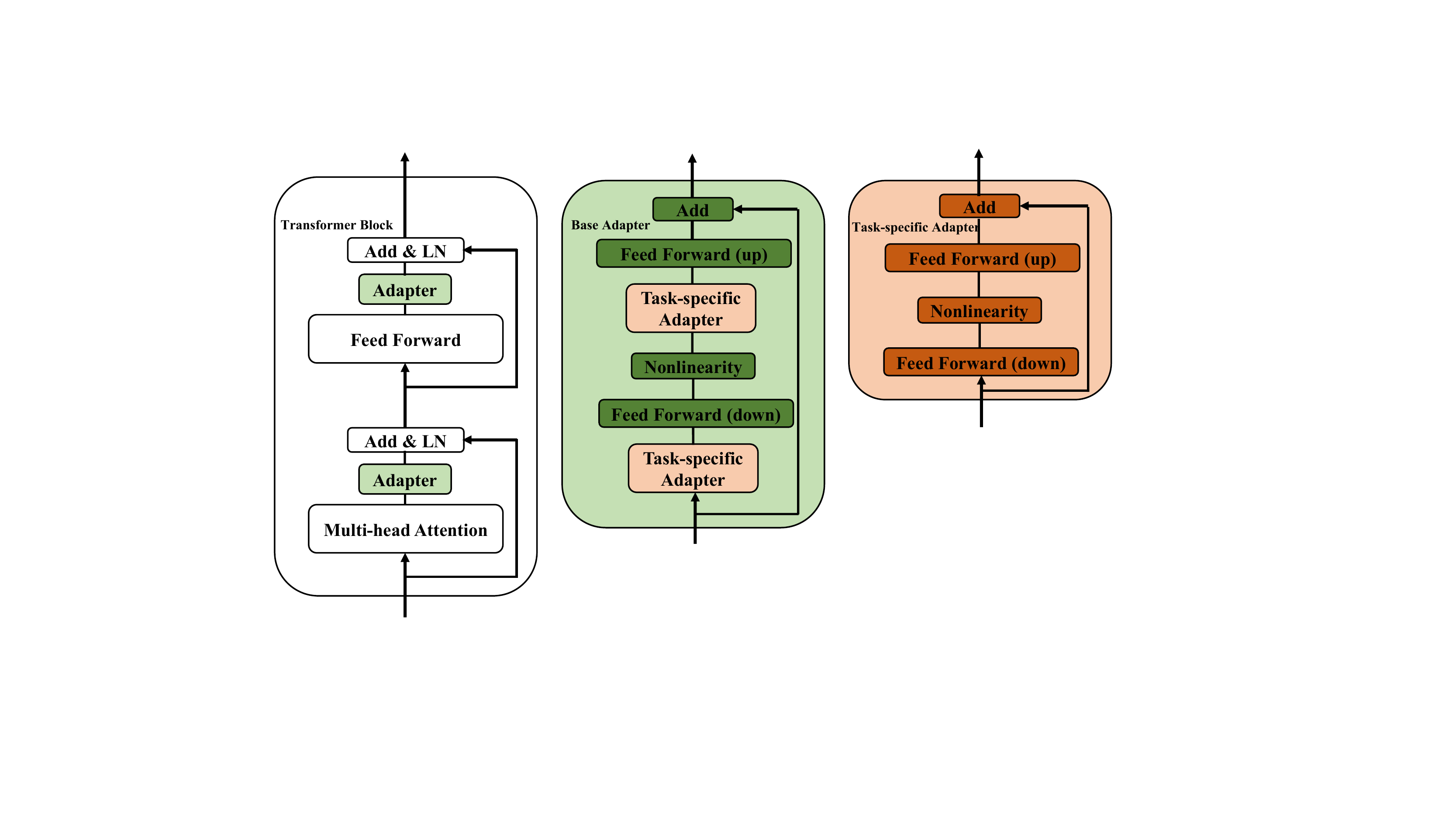}
    \caption{\textit{An overview of the hierarchical inductive transfer framework.}}
    \label{fig: framework}
\end{figure}

Directly fine-tuning pre-trained models for each of the new dialogue tasks will cause knowledge forgetting, and therefore each task requires a large set of parameters for maintaining the model performance on both old and new tasks. 
Compared with it, we keep the parameters of the pre-trained model fixed and use the adapter to learn new tasks. 
Adapters are inserted after the self-attention layer and the feed-forward layer of each block of the pre-trained model, illustrated in Figure \ref{fig: framework}:
\begin{equation}
\small
\mathbf{h}^{l+1} = 
\mathrm{LN}\left( 
\mathbf{h}^{l} + \mathrm{Ada}\left(\mathrm{Fun}\left(\mathbf{h}^{l}\right)\right)  
\right),
\end{equation}
where $\mathbf{h}^{l}$ and $\mathbf{h}^{l+1}$ represent the input and the output of sub-blocks, and $\mathrm{Fun(\cdot)}$, $\mathrm{Ada(\cdot)}$, and $\mathrm{LN(\cdot)}$ represent the function layer (i.e., the self-attention layer or the feed-forward layer), the adapter, and the layer norm, respectively.

Each adapter consists of a bottleneck module with a skip-connection. Concretely, the bottleneck module first down-projects the $d_{o}$-dimension output of the previous layer into features with a smaller dimension, $d_{a}$, followed by a nonlinearity, and then up-projects to the original dimension. Formally, it can be expressed as: 
\begin{equation}
\small
\mathrm{Ada}(\mathbf{o}) = \mathbf{o} + W^{U} a\left(W^{D} \mathbf{o}\right),
\end{equation} 
where $W^{D}$ ($d_{o} \times d_{a}$) and $W^{U}$ ($d_{a} \times d_{o}$) are the parameters of the down- and the up-projections, and $a(\cdot)$ is the activation function. By adjusting the value of $d_{a}$, we can control the number of parameters of adapters to a deployment-friendly range. 

For each new task, only a few parameters need to be trained on the cloud servers and delivered to the embedded devices, which significantly improves the training efficiency and reduces the size of the dialogue system. Please refer to Appendix \ref{ap:training efficiency of adapters} for a more detailed discussion.

\begin{table*}[t]
\centering
\small
\begin{tabular}{@{}l||c c||c c c| c||c@{}}
\hline
\bf Method & $\Theta$ & $\theta_\Delta$ & ConvAI2 & WoW & ED & BST & \bf Average \\ \hline
FE & + 0.216$\times$ & \phantom{00}5.4\phantom{0}\% & 0.8698 & 0.9129 & 0.6255 & 0.7413 & 0.7874 \\
FT & + 4.0\phantom{00}$\times$ & 100\phantom{.00}\% & 0.8855 & 0.917\phantom{0} & 0.6267 & 0.7838 & 0.8032 \\
MT+FT & + 2.0\phantom{00}$\times$ & 100\phantom{.00}\% & 0.8878 & \bf 0.9274 & 0.6241 & \bf 0.8241 & \bf 0.8158 \\ 
Ada & + 0.075$\times$ & \phantom{00}1.87\% & 0.888\phantom{0} & 0.9177 & 0.6204 & 0.7662 & 0.7981 \\\hline
AdaHIT & + 0.112$\times$ & \phantom{00}4.2\phantom{0}\% & \bf 0.8914 & 0.9193 & \bf 0.6358 & 0.8167 & \bf 0.8158 \\ \hline
\end{tabular}
\caption{Comparison in terms of total number of additional parameters ($\Theta$), trainable parameters per task ($\theta_\Delta$), and performance on tasks. The proposed AdaHIT achieves performance competitive with the state-of-the-art (MT+FT) with far fewer total parameters to be stored and parameters to be trained.}
\label{tab: automatic evaluation}
\end{table*}

\subsection{Hierarchical Inductive Transfer}
In continual dialogue learning, the old tasks usually contain useful knowledge for the learning of new tasks. But they may also have knowledge interference with new tasks. 
To alleviate this issue, one can multi-task the adapters with all old tasks and find general knowledge for new tasks. However, the regularization between diverse tasks also causes the performance decline of multi-tasked models on some old tasks due to knowledge interference among old tasks. 
Therefore, we further design a hierarchical inductive transfer framework that consists of two kinds of adapter, the base adapter and the task-specific adapter. 

Specifically, we take the vanilla adapters as the base adapters and introduce a set of new adapters inserted before the feed-forward layers of each base adapter as the task-specific sub-adapters, shown in Figure \ref{fig: framework}. It can be formulated as: 
\begin{equation}
\small
\mathrm{Ada_{bs}}(\mathbf{o}) = \mathbf{o} + W^{U} 
\mathrm{Ada_{ts}}\left(
a\left(W^{D} 
\mathrm{Ada_{ts}}\left(
\mathbf{o}
\right)\right)\right),
\end{equation} 
where $\mathrm{Ada_{bs}}(\cdot)$ and $\mathrm{Ada_{ts}(\cdot)}$ represent the base adapter and the task-specific adapter. Each task-specific adapter also consists of a bottleneck module and a skip-connection. 

During training, we first multi-task the base adapters with all old tasks to find general knowledge and then fine-tune a set of task-specific adapters for each task, including old tasks and new tasks, which enables the new task to benefit from the knowledge of old tasks without sacrificing the model performance on some old tasks.
\section{Experiment}

\subsection{Datasets and Baselines}
\paragraph{Datasets} To evaluate the proposed framework, we take \textbf{ConvAI2} (an extension of the PersonaChat dataset \cite{DBLP:conf/acl/KielaWZDUS18}), Wizard of Wikipedia (\textbf{WoW}) \cite{DBLP:conf/iclr/DinanRSFAW19}, Empathetic Dialogues (\textbf{ED}) \cite{DBLP:conf/acl/RashkinSLB19}, and Blended Skill Talk (\textbf{BST}) \cite{DBLP:conf/acl/SmithWSWB20} as an instance of continual dialogue learning. The first three tasks are the old tasks and the last task represents the new task.

\paragraph{Baselines} Four methods of inductive transfer are used to compare with our framework (\textbf{AdaHIT}), including feature extraction (\textbf{FE}), which adds and optimizes a classification layer on the top of the pre-trained model \cite{DBLP:conf/nips/VaswaniSPUJGKP17}, fine-tuning (\textbf{FT}), which updates all parameters of the pre-trained model for each task, multi-tasking with fine-tuning (\textbf{MT+FT}), which first multi-tasks the entire pre-trained model with all old tasks and then fine-tunes it on the new task, and vanilla adapter (\textbf{Ada}), which trains a set of adapters for each task.

\subsection{Experimental Settings}
Following \citet{DBLP:conf/acl/SmithWSWB20}, we use the poly-encoder with 256M parameters \cite{DBLP:journals/corr/abs-1905-01969} as the underlying model, pretrain it on the pushshift.io Reddit dataset, and then conduct inductive transfer on the downstream tasks. 
We also truncate the length of label and text to 72 and 360, and set the embedding size to 768 as \citet{DBLP:conf/acl/SmithWSWB20}. 
The batch size is 128 and the other responses in a batch are set as negatives for training. 
The dimension of adapters $d_{a}$ is 64. 
We adopt AdaMax \cite{DBLP:journals/corr/KingmaB14} as the optimizer throughout the experiments, and the learning rates are 9e-4, 2.5e-3, 1e-3, and 4e-4 for ConvAI2, WoW, ED, and BST. 
The total training epochs are 8 with linear warm-up for 10\% and linear decay for the rest. 
All experiments are conducted using ParlAI\footnote{\url{https://parl.ai/}}.

\subsection{Experimental Results}
For the retrieval-based dialogue scenarios, we measure hits@1/K\footnote{hits@1/K represents recall@1 when choosing the gold response from K candidates.} on the validation set of each task for automatic evaluation. The number of candidates is 20 for ConvAI2 and 100 for other tasks. 
The results reported in Table \ref{tab: automatic evaluation} show that AdaHIT achieves the best average performance, the same as MT+FT, at the cost of far fewer parameters to be trained and stored, indicating the superiority of deployment on embedded devices. 
AdaHIT significantly outperforms Ada in both old tasks and new task with a slight regression of computational efficiency, which demonstrates that the hierarchical inductive transfer can extract general knowledge to facilitate the learning of the new task while boosting the model performance on old tasks effectively.

\subsection{Ablation Study and Analysis}
\paragraph{Effect of Base Adapter}
To analyze the effect of the base adapter, we train it with different tasks, and then test it on BST in a zero-shot manner, or a fine-tuning manner which is the same with AdaHIT. 
From the results in Table \ref{tab: Impact of Base Adapter}, we can observe that the base adapter with multi-tasking obtains the best performance under both the zero-shot and the fine-tuning setting, indicating that multi-tasking provides more general knowledge for the learning of BST. 
It is also worth mentioning that the base adapter trained on ConvAI2 achieves better performance than adapters on other tasks, because ConvAI2 contains more useful information, e.g., persona, that also exists every sample of BST. 

\begin{table}[t]
\centering
\small
\setlength{\tabcolsep}{2.5pt}

\begin{tabular}{@{}l|cc@{}}
\hline
\bf Dataset for Adapter & \multicolumn{1}{c}{BST (Zero-Shot)} & BST (Fine-Tuning) \\ \hline
ConvAI2 & 0.753\phantom{0} & 0.8039 \\
WoW & 0.6222 & 0.7751 \\
ED & 0.6349 & 0.7846 \\ \hline
MT & \bf0.768\phantom{0} & \bf0.8167 \\ \hline
\end{tabular}
\caption{Effect of training datasets for the base adapter.}
\label{tab: Impact of Base Adapter}
\end{table}

\begin{figure}[t]
\centering
\includegraphics[width=0.90\linewidth]{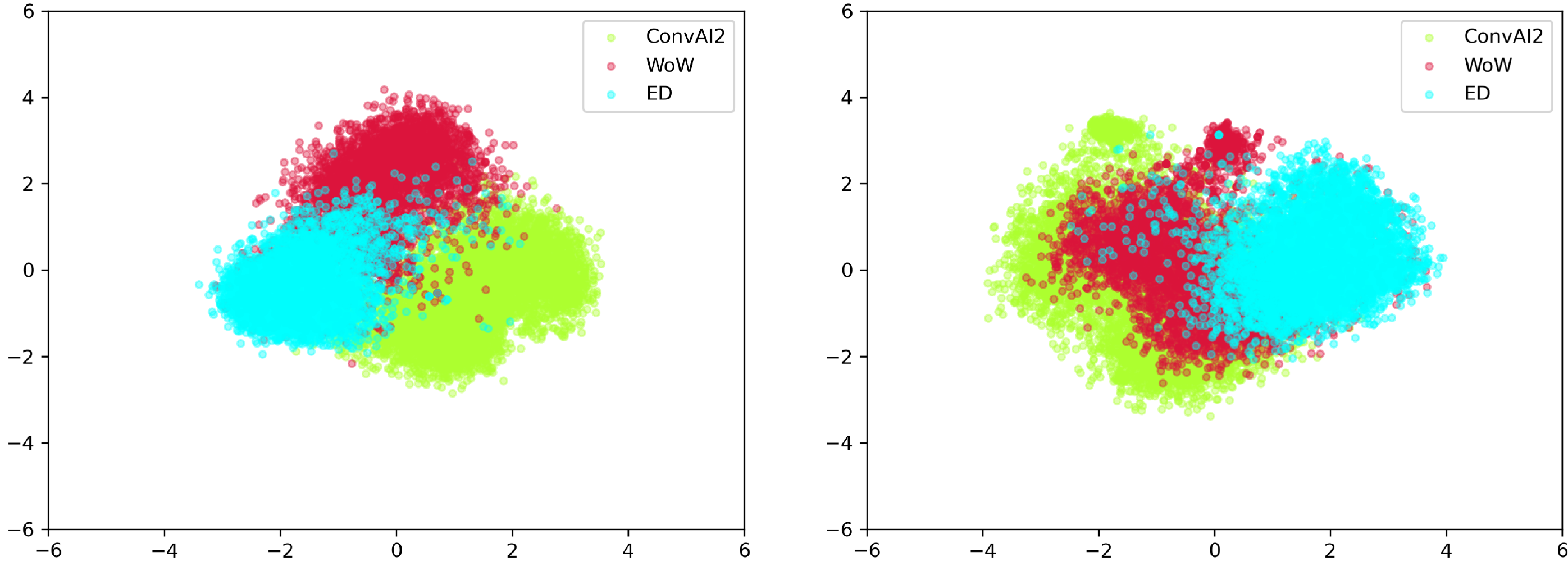}
\caption{Visualization of learned sentence representations from AdaHIT with differently-trained base adapters. MT is on the left and ConvAI2 is on the right.}
\label{fig: visualization}
\end{figure}

\paragraph{Visualization}
To verify whether AdaHIT helps task adaption, we visualize the representations from models with different base adapters, i.e., trained on MT and ConvAI2, the result of which is shown in Figure~\ref{fig: visualization}. As we can see, the two models can both adjust to specific downstream tasks but representations with MT are better distributed and more tightly clustered. It is also interesting to see that the model with MT may implicitly distinguish the skills for each task, because while ED and ConvAI2 share more common skills, they are quite different from WoW, and such difference is evidently reflected by the visualization.

\begin{table}[t]
\centering
\small
\setlength{\tabcolsep}{2pt}

\begin{tabular}{@{}l|llllll@{}}
\hline
\bf Number of Layers & \multicolumn{1}{c}{1} & \multicolumn{1}{c}{2} & \multicolumn{1}{c}{3} & \multicolumn{1}{c}{4} & \multicolumn{1}{c}{5} & \multicolumn{1}{c}{6} \\ \hline
AdaHIT & 0.809 & 0.807 & 0.796 & 0.785 & 0.762 & 0.734 \\ \hline\hline
\bf Position of Layers & \multicolumn{1}{c}{0} & \multicolumn{1}{c}{2} & \multicolumn{1}{c}{4} & \multicolumn{1}{c}{6} & \multicolumn{1}{c}{8} & \multicolumn{1}{c}{10} \\ \hline
AdaHIT & 0.809 & 0.808 & 0.809 & 0.801 & 0.793 & 0.763 \\ \hline
\end{tabular}
\caption{Ablation Study in terms of number and position of removed adapters on BST.}
\label{tab: ablation study}
\end{table}

\paragraph{Ablation Study}

We further investigate the impact of adapters on model performance quantitatively. First, we gradually remove each trained adapter from the bottom layer, and then increase the number of removed adapters. As shown in Table \ref{tab: ablation study}, the adapters of higher layers have more significant effects than the adapters of lower layers, indicating that we can only insert the adapters into the higher layers to improve the training efficiency.
\section{Related Work}

\paragraph{Continual Dialogue Learning} 
Neural dialogue models \cite{DBLP:conf/coling/MouSYL0J16,DBLP:conf/aaai/XingWWLHZM17,DBLP:conf/acl/ZhaoZE17,DBLP:conf/aaai/ShenSND18,DBLP:conf/aaai/FengRL021} can acquire various kinds of conversation skills from corpora, such as characterizing personalities \citep{DBLP:conf/ijcai/QianHZXZ18,DBLP:conf/acl/KielaWZDUS18}, expressing emotion and empathy \citep{DBLP:conf/aaai/ZhouHZZL18,DBLP:conf/acl/RashkinSLB19}, and retrieving knowledge \citep{DBLP:conf/aaai/GhazvininejadBC18,DBLP:conf/iclr/DinanRSFAW19}. Unlike existing work on enhancing a particular conversation skill, we work towards a new dialogue learning paradigm, where conversation skills are gradually embedded into a single model by mutual reinforcement instead of interference.

\paragraph{Inductive Transfer} 
Continual learning in terms of transferring inductive knowledge from pre-trained models to downstream tasks can be categorized into feature-based, fine-tuning--based, and adapter-based \citep{Ruder2019NeuralTL}. 
We adopt the adapter-based approach that benefits from both pre-trained models and a small set of extra parameters for task-specific knowledge. 
Unlike conventional adapters \cite{DBLP:conf/icml/HoulsbyGJMLGAG19,DBLP:conf/emnlp/PothPRG21,DBLP:conf/eacl/PfeifferKRCG21}, knowledge in the proposed adapters will be used to further boost the learning of new dialogue tasks, whereas knowledge of each task is separated into general and task-specific parts to avoid knowledge interference. 
\citet{DBLP:journals/corr/abs-2008-12579} also uses the adapters to acquire the conversation skills, but it does not consider knowledge transfer and interference between adapters. 
\section{Conclusion}
In this work, we propose a hierarchical inductive transfer framework to efficiently train and deploy the pre-trained models for growing numbers of new dialogue tasks requiring diverse skills. 
Considering the computation resource--limited embedded devices, we first adopt the adapter module, a small plug-in sub-net, as the only incremental and trainable parameters for learning each of the new dialogue tasks. 
To take advantage of knowledge in old tasks to facilitate the learning of new tasks, we further propose the hierarchical inductive transfer to alleviate knowledge interference between tasks and provide general knowledge for new tasks. 
Extensive experiments and analysis demonstrate that the proposed framework achieves high computational efficiency with competitive performance.

\section*{Acknowledgements}
This research is supported by Beijing Natural Science Foundation (No. 4222037 and L181010), National Natural Science Foundation of China (No. 61972035), Natural Science Foundation of China (NSFC) No. 62176002, and Beijing Academy of Artificial Intelligence (BAAI). Xu Sun and Kan Li are the corresponding authors.

\bibliographystyle{acl_natbib}
\bibliography{anthology,acl2022}

\appendix

\section{Structure of Adapters}\label{ap:structure of adapters} 
We have designed and evaluated diverse structures of adapters for continual dialogue tasks, such as the self-attention structure and the convolutional structure. However, there is no significant effect on performance, which is in line with previous adapter-based work. 
For the basic bottleneck structure, there are two advantages. First, it can limit the number of parameters per adapter by setting the bottleneck dimension $d_a \ll d_o$. Second, it also provides a flexible way to trade-off model performance with parameter efficiency.

\begin{table}[h]
\centering
\small

\begin{tabular}{l|lll}
\hline
Method & ConvAI2 & WoW & ED \\ \hline
MT (B-FT) & 0.8878 & 0.9274 & 0.6241 \\ \hline
MT (A-FT) & 0.8767 & 0.9094 & 0.6136 \\ \hline
\end{tabular}
\caption{Results on the old tasks. MT (B-FT) and MT (A-FT) represent the multi-tasking model before and after being fine-tuned on the new task, respectively}
\label{tab:knowledge forgetting of ft}
\end{table}
\section{Training Efficiency of Adapters}\label{ap:training efficiency of adapters} 
Compared with the traditional fine-tuning method, our framework conducts the learning of dialogue tasks only by adapters, which reduces the memory requirements and the computing operations of each batch and therefore trains more samples with the same time. 
For example, there is a two-layer network, and only the first layer is trainable: 
$$
y_1 = f(w_1 *x + b_1)
$$
$$
y_2 = f(w_2 *y_1 + b_2)
$$
Although we still need to calculate $\frac{\partial y_2}{\partial y_1}$ due to the chain rule, we do not calculate $\frac{\partial y_2}{\partial w_2}$ and $\frac{\partial y_2}{\partial b_2}$ (i.e., reducing the computing operations) and do not save them (i.e., reducing the memory requirements) for the parameter update. 
Considering the number of parameters of Transformer, the proposed framework indeed improves the training efficiency. 
Moreover, we can only insert the adapters in the top layers because the adapters in the bottom layers have a weaker effect on the model performance, indicated by Table \ref{tab: ablation study}, which limits the chain derivative to the top layers and further reduces the computing operations.

\section{Knowledge Forgetting of FT}\label{ap:knowledge forgetting of ft} 
In order to demonstrate knowledge forgetting of the traditional fine-tuning method, we evaluate the performance of the multi-tasking model (MT) on the old tasks before and after being fine-tuned on the new task. As shown in Table \ref{tab:knowledge forgetting of ft}, the model fine-tuned on the new task (BST) shows consistent performance degradation on the old tasks.

\begin{table}[t]
\centering
\small

\begin{tabular}{llll|l}
\hline
ConvAI2 & WoW & ED & BST & Average \\ \hline
0.8833 & 0.9233 & 0.6288 & 0.8342 & 0.8174 \\ \hline
\end{tabular}
\caption{Results of the model that is first pre-trained on the old tasks and then multi-tasked on all tasks.}
\label{tab:multi-tasking on all tasks}
\end{table}

For a fair comparison, both our method (AdaHIT) and MT+FT are multi-tasked on the old tasks and then fine-tuned on the new task. We also provide the results of a stronger model that is first pre-trained on the old tasks and then multi-tasked on all tasks (i.e., both the old and the new tasks). The results of Table \ref{tab:multi-tasking on all tasks} show that AdaHIT still achieves comparable performance but consumes less computational cost.

\end{document}